# Design and Biomechanical Evaluation of a Lightweight Low-Complexity Soft Bilateral Ankle Exoskeleton

Josée Mallah, Zakii Javed, Zafer Azak, Thomas Stone, and Luigi G. Occhipinti, *Senior Member, IEEE*

*Abstract—* Many people could benefit from exoskeleton assistance during gait, for either medical or nonmedical purposes. But exoskeletons bring added mass and structure, which in turn require compensating for. In this work, we present a lightweight, low-complexity, soft bilateral ankle exoskeleton for plantarflexion assistance, with a shoe attachment design that can be mounted on top of any pair of shoes. Experimental tests show no significant difference in lower limb kinematics and kinetics when wearing the exoskeleton in zero-torque mode relative to not wearing an exoskeleton, showing that our device does not obstruct the wearer's healthy gait, and proving it as a compliant and comfortable device, promising to provide effective assistance. Hence, a control system was developed, and additional tests are underway.

## I. INTRODUCTION

Walking is a basic locomotion skill and consists in most of human daily movement [1], [2]. While it is extensively used by many healthy individuals to undertake physically demanding tasks, such as walking on rough terrain and carrying heavy weights (e.g. soldiers), many other people suffer from medical conditions that hinder their daily living and autonomy, such as spinal cord injury, neurological disorders, trauma (e.g. stroke), cerebral palsy, etc [3]. However, both categories could benefit from wearable robotic assistance in the form of exoskeletons, which have been developed for both medical and nonmedical purposes. Medical exoskeletons help restore limited or lost body functions due to medical conditions and are used for rehabilitation purposes under professional supervision. Nonmedical devices, on the other hand, usually come as augmentation devices, assisting healthy people through strength supplementation, speed augmentation, or reduction of effort and metabolic cost required for some activities; exoskeletons developed for elderly assistance in the context of healthy aging also come under this category [3].

In particular, ankle exoskeletons have garnered significant attention, as the ankle joint plays an essential role in human gait, producing more mechanical power than either the knee or the hip, and supporting balance during walking [4], [5], [6]. It is particularly affected by specific conditions, such as stroke and aging, which makes it responsible for the limited mobility of the individuals suffering from those conditions [4], [5], [6]. In particular, weakness in plantarflexor muscles affects stability and propulsion [4], [7]. Assisting the ankle joint can provide optimal metabolic consumption savings [8], as increasing the mechanical power output of the ankle decreases the power demands on the hip, leading to improved walking efficiency [6].

Two types of exoskeletons have been developed over time, from an actuation point of view: passive and active. Despite their light weight, passive devices have limited controllability and struggle to break the metabolic cost barrier, due to their inability to provide enough positive power to overcome the added device mass [9], [10], [11], and therefore, their active counterparts were proven more effective [3]. While many platform exoskeletons have been proposed, being beneficial for keeping the actuator mass off the wearer's body, wearable untethered exoskeletons can be worn during daily life, providing assistance on the go [3]. Hence, active wearable designs make the optimal choice.

Exoskeletons can be classified into rigid, hybrid, and soft based on the rigidity of the structure used. Hybrid and soft designs often opt for flexible force transmission, using Bowden cables [12], [13], [14] and present an advantage in terms of compliance and user comfort, which makes the devices ideal for extended use. Many exoskeletons provide unilateral assistance, which is beneficial in cases such as stroke, where only one side is impaired; however, for other user categories, especially healthy users with no impairments, bilateral devices would preserve gait symmetry, and are known to provide greater metabolic benefits [15].

One problem with exoskeletons is the added mass and rigid structure, as increasing exoskeleton weight impedes gait (increased step length, decreased step height [16]) and dynamic balance [17] and increases metabolic consumption, especially when the weight is added distally on the legs [18], [19], even shifting the focus of the assistance to merely offsetting the added device mass and keeping the same metabolic consumption levels as during unassisted movement [20], [21]. In fact, many studies have reported changes in kinematics and/or kinetics away from the desired healthy walking patterns in zero-torque mode [13], [22], [23]. Hence the interest in softer, lighter devices [3].

Moreover, we recently proposed a complexity index in [3] allowing to evaluate how complex an exoskeleton design is, with an obvious preference for less complex devices, which are inherently easier to develop and less expensive. The complexity index is defined as a weighed sum of factors spanning the number of assisted limbs (L), the number of degrees of freedom (D), and the numbers of sensors (S) and

*J. Mallah is funded by an Allen, Meek, and Read Cambridge International Scholarship from Cambridge Trust. This study was supported by MathWorks–Cambridge University Engineering Department (CUED) Small Grants Programme 2025.

J. Mallah and L. G. Occhipinti[*] are with the Electrical Engineering Division, Department of Engineering, University of Cambridge, Cambridge, CB3 0FA, United Kingdom ([*] corresponding author: phone: 01223 3 32838; e-mail: lgo23@cam.ac.uk).

Z. Javed, Z. Azak, and T. Stone are with the Cambridge University Hospitals NHS Foundation Trust, Cambridge CB22 5LD, United Kingdom.

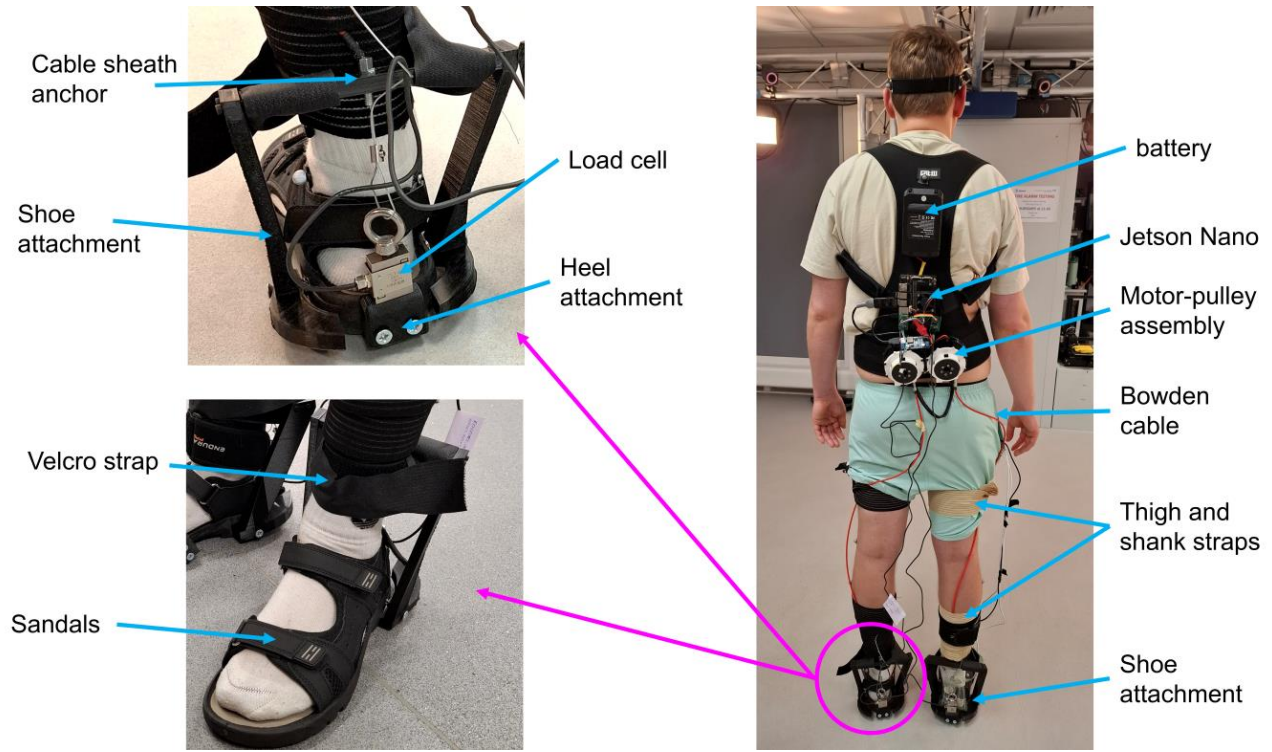

Figure 1. Full exoskeleton worn by the study participant, comprising the back support structure carrying the battery, the controller, and the actuation assembly. The Bowden cables run all the way down to the ankle, and are routed on the legs using textile straps at the thighs and shanks. The shoe attachment, including the cable sheath anchoring structure and heel attachment, is shown closely on the left, along with the Velcro strap that secures it to the shank.

actuators (A) in a device. Therefore, we are aiming for a device with low complexity.

This paper introduces a lightweight, low-complexity, active bilateral ankle exoskeleton with hybrid rigidity (Fig. 1), based on a shoe attachment on top of flexible, comfortable shoes, and observes its impact on lower limb biomechanics in zero-torque mode, proving its compliance, being non-obstructing to the wearer's gait. We also discuss the control system design, with plans to test the powered exoskeleton in future work.

## II. ELECTROMECHANICAL DESIGN OF THE ANKLE EXOSKELETON

We developed a wearable, hybrid-rigidity, battery-powered, and self-contained ankle exoskeleton capable of providing bilateral plantarflexion assistance. A back-mounted actuation assembly transmits force through Bowden cables to the ankle joint: the inner Bowden cable runs all the way down to a load cell mounted on a heel attachment, while the external sheath is anchored near the lower shank. We wanted to keep the design "as soft as possible"; however, the Bowden sheath anchor at the shank experiences significant downwards forces when the inner cable is pulled up, a problem which was explained by [24], and which could potentially explain why fully soft ankle exoskeletons are still a minority [3]. Hence, we designed a rigid sheath anchor (Fig. 2) that mounts on the shoe, preventing the sheath from sliding downwards, and allowing for effective force transmission. The vertical rods go up on the sides away from the user's leg, and the horizontal bar goes behind the shank, and is wrapped around it using a Velcro strap, which allows to fit different calf sizes, makes sure that the rigid structure barely touches the wearer's leg, and does not obstruct its natural movement. This structure comes as a shoe attachment instead of a full foot module or shoe replacement; it can be mounted onto any pair of shoes, which could contribute to increased user acceptance. For demonstration, we are using a pair of strapped sandals, which fits people with neighboring shoe sizes.

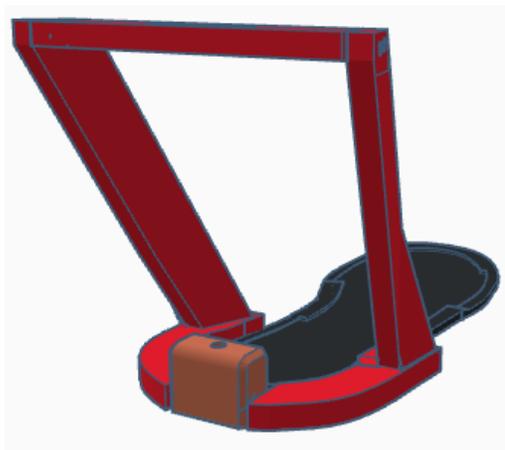

Figure 2. CAD drawing, showing the cable sheath anchoring structure (red) and the heel attachment serving as load cell mount (brown).

The actuation assembly consists of two quasi-direct-drive motors (PDA08, Daran Robotics), each connected to a pulley and a Bowden cable. The inner Bowden cable is connected to a 500 N load cell (DYLY08, DYSENSOR) mounted on a heel attachment, allowing to measure the transmitted tension and provide feedback to the control system to ensure proper tracking of the desired torque. The back-mounted assembly also includes the processor board (Jetson Nano B01, NVIDIA), which communicates with the motors via CAN bus interface, and a 24 V, 4 Ah lithium-ion battery pack (G24B2, ASUNCELL) which powers the motors and the processor via a 24-5 V step-down converter. Force sensitive resistors (FSR402, Youmile) are placed on the shoe sole beneath the heel, allowing for heel strike detection. The overall device mass remains below 2.5 kg (Table I).

## III. EXPERIMENTAL SETUP

### A. Data Collection

This study was approved by the Ethics Committee at the Department of Engineering, University of Cambridge (Reference No. 639). One healthy subject was recruited (29-year-old male, 1.78 m, 80.4 kg). The same subject had been participating in multiple exoskeleton testing sessions and has therefore acquired significant experience walking with the device. The data were collected in the gait lab at the Cambridge Prosthetic and Orthotic Service. Motion capture data were collected using a Vicon system (©Vicon Motion Systems Ltd., Oxford, UK) comprising 12 infrared cameras and two synchronized video cameras. Ground reaction forces were collected using two AMTI Hall-effect force plates (©Advanced Mechanical Technology, Inc., Watertown USA). A Helen Hayes marker set was applied using the Vicon Plug-in Gait model. Reflective markers were placed bilaterally on the anterior superior iliac spines (ASIS), posterior superior iliac spines (PSIS), lateral tibias/shanks, lateral thighs, lateral femoral epicondyles, lateral malleoli, heels (posterior calcaneus), and second metatarsal heads (forefoot/toe), in addition to a few major upper body landmarks. The participant walked in straight lines at a self-selected speed.

### B. Experimental Conditions

Two walking conditions were considered, allowing to observe the separate impact of the exoskeleton added mass: (1) No Exoskeleton – the participant walks in unmodified shoes, similar to those used in the exoskeleton; (2) Exoskeleton Off – the participant walks while wearing the exoskeleton unpowered/in zero-torque mode (Fig. 1).

TABLE I. EXOSKELETON WEIGHT BREAKDOWN

| Part | Weight (g) |
|---|---|
| Motors | 812 |
| Battery | 670 |
| Jetson Nano | 141 |
| Cables | 48 |
| Shoe attachments | 600 |
| **Total** | 2,271 (< 2,500 overall) |

### C. Data Processing

Kinematic and kinetic data were processed using the Vicon Nexus software (version 2.18) and saved to c3d files. A Woltring filter with a mean square error of 10 [25] was applied to the marker trajectories, as per the Vicon Nexus default configuration. Each trial was reconstructed to create 3D markers from raw camera data, the marker trajectories were labeled; pre-set gap-filling and filtering tools were used to fill the gaps in the marker trajectories. Gait events were manually marked upon visual observation of video recordings.

Python scripts were then used to extract gait events, kinematic, and kinetic data from the c3d files, using the ezc3d library [26]. Kinematics and kinetics were segmented by stride and time-normalized to 101 samples corresponding to 0-100% gait cycle using linear interpolation.

To compare between the two experimental conditions, a linear mixed-effects model (LME) and two one-sided tests (TOST) were used. For each scalar feature, an LME was fit at the stride level with condition as a fixed effect and trial as a random intercept, as each trial includes multiple strides. Then, trial-level means were computed over the strides, and TOST was performed using a Welch standard error, with equivalence bounds of ±2° for the angles and ±0.05s for the durations. Statistical significance was set to $\alpha = 0.05$.

## IV. RESULTS

In order to evaluate the biomechanical impact of the added exoskeleton mass and structure, we observe the kinematics the three major lower-limb joints: ankle, knee, and hip flexion, along with the kinetics of the ankle, and some temporal gait parameters.

### A. Kinematics

Fig. 3-a shows the ankle joint angles in the No Exoskeleton and Exoskeleton Off modes. Table II provides the average range of motion ROM of each joint in both conditions.

The statistical analysis shows no significant difference in ankle plantarflexion (LME: p=0.46, TOST: difference=0.2). A slight difference is observed for dorsiflexion (LME: p<0.01, diff=1.24°) but still within TOST equivalence bounds (diff=1.32°). There is no statistically significant difference in range of motion (p=0.15).

### B. Ankle Kinetics

Fig. 3-b shows the ankle flexion moment in both experimental conditions over the gait cycle.

The statistical analysis shows no difference in peak plantarflexion moment (p=0.98) between the two experimental conditions.

### C. Temporal Parameters

Table III presents the temporal parameters considered: gait cycle duration, stance and swing phases durations and percentages of the gait cycle.

Despite having a statistically significant difference in gait cycle duration (LME: p<0.01), the durations are equivalent at the ±0.05s bound (TOST: diff=-0.0241). Stance and swing times and percentages have no significant differences (p=0.39, 0.98, 0.51, 0.98 respectively).

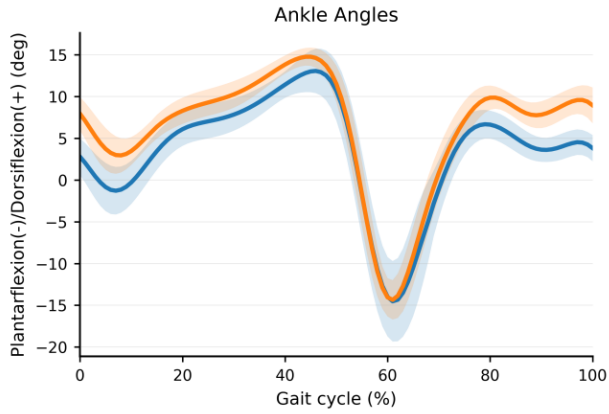

(a)

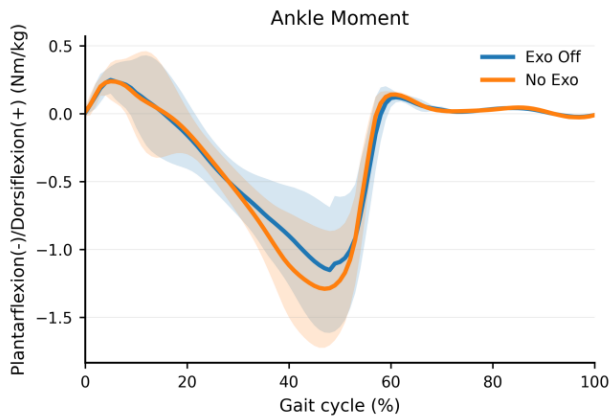

(b)

Figure 3. Ankle kinematics (a) and kinetics (b) in No Exoskeleton (orange) and Exoskeleton Off (blue) modes.

TABLE II. AVERAGE JOINT RANGE OF MOTION WITHOUT AND WITH EXOSKELETON

| Joint | ROM (°) | |
|---|---|---|
| | No Exoskeleton | Exoskeleton Off |
| Ankle Flexion | 29.785 ± 1.907 | 29.096 ± 4.017 |
| Knee Flexion | 55.976 ± 3.525 | 59.124 ± 5.528 |
| Hip Flexion | 47.939 ± 2.217 | 45.994 ± 5.848 |

TABLE III. TEMPORAL GAIT PARAMETERS WITHOUT AND WITH EXOSKELETON

| Parameter | No Exoskeleton | Exoskeleton Off |
|---|---|---|
| Gait cycle duration (s) | 0.980 ± 0.027 | 1.003 ± 0.032 |
| Stance phase duration (s) | 0.559 ±0.023 | 0.572 ± 0.108 |
| Stance phase percentage (GC%) | 57.072 ± 1.795 | 56.934 ± 10.356 |
| Swing phase duration (s) | 0.421 ± 0.022 | 0.4311 ± 0.098 |
| Swing phase percentage (GC%) | 42.928 ± 1.795 | 43.066 ± 10.356 |

## V. CONTROL DESIGN

The control algorithm consists of a closed loop torque/cable tension control (Fig. 4). A proportional integral derivative (PID) controller with a feed forward element is used to track the desired cable tension, which is tracked by the load cell. The torque was defined as a timed trajectory based on Zhang et al. [27], using four main parameters: rise time, peak time, fall time (gait cycle percentage GC%), and peak torque (Nm). In this paper, we set the parameters to 23.2, 50.4, 62.7 GC%, and 17 kg of tension, which is equivalent to 10 Nm at the ankle. Heel strike detection was achieved using FSRs, allowing to compute GC%.

## VI. DISCUSSION

The objective of this paper was to introduce a lightweight hybrid-rigidity bilateral ankle exoskeleton and evaluate its impact on lower limb biomechanics in zero torque mode. A control system was designed and is planned to be tested in the gait lab in future work.

We aimed for a relatively soft exoskeleton design, for the parts in contact with the user's body, using a flexible transmission and textile straps. A rigid cable anchoring structure ended up being crucial to ensure effective force transmission to the ankle but is presented as a shoe attachment that can be mounted on the user's preferred pair of shoes.

These design choices allowed for a lightweight final device (<2.5 kg), which comes in the very lower end for bilateral ankle exoskeletons, for which the weight range [3] goes from around 1.73-2.07 kg for Orekhov et al. [12] up to 10 kg in Li et al. [28] and is well below the 4 kg limit recommended by [3] as an optimal compromise between metabolic savings, design complexity, and device weight.

Furthermore, in terms of design complexity, our present exoskeleton achieves a complexity index of 2 (L = 2, D = 1, S = 4, A = 2), placing it again on the lower end for bilateral ankle exoskeletons, relative to a range [3] from 1.7 for both Orekhov et al. [29] and Mooney et al. [30] to 4.4 for Zhao et al. [31].

Lower-limb joints – ankle, knee, and hip – are closely related, and altering the behavior of one is likely to affect the others. Moreover, many exoskeleton studies have reported unnatural gait patterns in zero-torque mode. Gasparri et al. [22] reported excessive ankle dorsiflexion and a reduced ankle ROM. Xia et al. [13] also had a general trend of reduced ankle ROM across subjects. Shafer et al. [23] reported excessive peak biological ankle moments. Therefore, we evaluated the impact of our device in zero-torque mode on the kinematics of the three joints, as well as on the kinetics of the ankle, to ensure that the user's ankle joint does not need to provide additional biological torque to counteract the exoskeleton added mass and structure. Despite seeing a difference in peak knee flexion between the two experimental conditions in Table II, LME verifies that the difference is not significant (p=0.24). Therefore the experimental results show that the kinematics of the three lower-limb joints, the kinetics of the ankle, and the temporal parameters of gait, in terms of the total gait cycle duration and stance and swing duration and percentages, do not generally have significant difference between the No Exoskeleton and Exoskeleton Off conditions, proving that our exoskeleton does not obstruct a healthy user's gait.

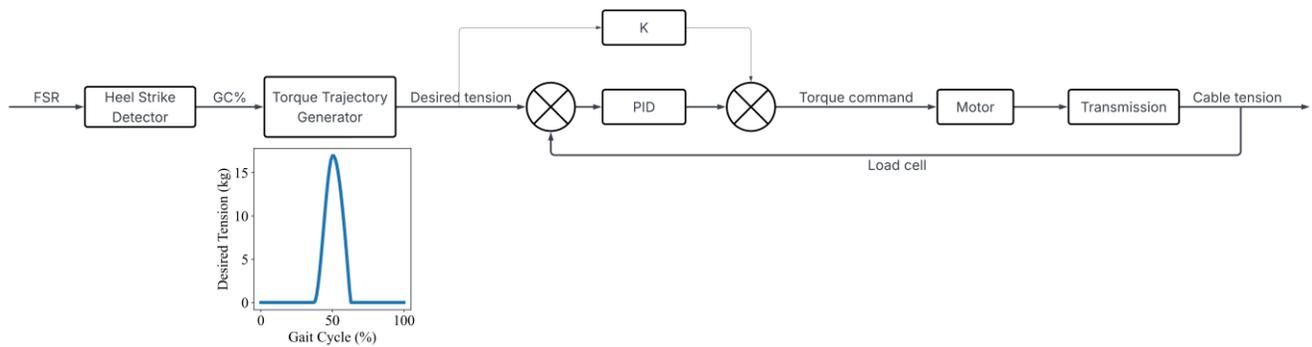

Figure 4. Control system architecture

Eventually, we propose a control system for the exoskeleton, based on PID tension control at the low-level, and a GC% estimation based on heel strike detection from FSRs, allowing to track a desired timed tension/torque trajectory at the mid-level. Future work would evaluate the powered exoskeleton, as well as reconsider the mechanical transmission design in a way to potentially allow for the provision of higher torques and expand the design to provide dorsiflexion assistance.

ACKNOWLEDGMENT

The authors would like to acknowledge advice and support from Z. Wang and F. Forni in control system design, as well as Y. Zhu and O. Griffiths in mechanical system design and prototyping.

REFERENCES

[1] H. G. Chambers and D. H. Sutherland, 'A Practical Guide to Gait Analysis', *JAAOS - J. Am. Acad. Orthop. Surg.*, vol. 10, no. 3, 2002, [Online]. Available: https://journals.lww.com/jaaos/fulltext/2002/05000/a_practical_guide_to_gait_analysis.9.aspx

[2] J. Mallah, Y. Zhu, K. Xu, G. S. Virk, S. Bai, and L. G. Occhipinti, 'Real-Time Prediction of Lower Limb Joint Kinematics, Kinetics, and Ground Reaction Force using Wearable Sensors and Machine Learning', Jan. 2026, doi: 10.48550/arXiv.2601.18494.

[3] J. Mallah *et al.*, 'A Systematic Review of Sensing and Control Strategies of Wearable Active Ankle Exoskeletons/Exosuits, With Performance Assessment Based on a Complexity Index', *IEEE Sens. Rev.*, vol. 2, pp. 382–396, 2025, doi: 10.1109/SR.2025.3588140.

[4] B. Shi *et al.*, 'Wearable Ankle Robots in Post-stroke Rehabilitation of Gait: A Systematic Review', *Front Neurorobot*, vol. 13, p. 63, 2019, doi: 10.3389/fnbot.2019.00063.

[5] K. E. Gordon and D. P. Ferris, 'Learning to walk with a robotic ankle exoskeleton', *J Biomech*, vol. 40, no. 12, pp. 2636–44, 2007, doi: 10.1016/j.jbiomech.2006.12.006.

[6] S. N. Fickey, M. G. Browne, and J. R. Franz, 'Biomechanical effects of augmented ankle power output during human walking', *J Exp Biol*, vol. 221, no. Pt 22, Nov. 2018, doi: 10.1242/jeb.182113.

[7] E. A. Morris *et al.*, 'Actuation Timing Strategies for a Portable Powered Ankle Foot Orthosis', presented at the ASME 2011 Dynamic Systems and Control Conference and Bath/ASME Symposium on Fluid Power and Motion Control, 2011, pp. 807–814. doi: 10.1115/dscc2011-6170.

[8] P. W. Franks, G. M. Bryan, R. M. Martin, R. Reyes, A. C. Lakmazaheri, and S. H. Collins, 'Comparing optimized exoskeleton assistance of the hip, knee, and ankle in single and multi-joint configurations', *Wearable Technol*, vol. 2, p. e16, 2021, doi: 10.1017/wtc.2021.14.

[9] S. Viteckova *et al.*, 'Empowering lower limbs exoskeletons: state-of-the-art', *Robotica*, vol. 36, no. 11, pp. 1743–1756, 2018, doi: 10.1017/s0263574718000693.

[10] L. Quinto, P. Pinheiro, S. B. Goncalves, I. Roupa, P. Simões, and M. Tavares da Silva, 'Analysis of a passive ankle exoskeleton for reduction of metabolic costs during walking', *Def. Technol.*, vol. 37, pp. 62–68, Jul. 2024, doi: https://doi.org/10.1016/j.dt.2023.11.015.

[11] E. Etenzi, R. Borzuola, and A. M. Grabowski, 'Passive-elastic knee-ankle exoskeleton reduces the metabolic cost of walking', *J. NeuroEngineering Rehabil.*, vol. 17, no. 1, p. 104, Jul. 2020, doi: 10.1186/s12984-020-00719-w.

[12] G. Orekhov, Y. Fang, J. Luque, and Z. F. Lerner, 'Ankle Exoskeleton Assistance Can Improve Over-Ground Walking Economy in Individuals With Cerebral Palsy', *IEEE Trans. Neural Syst. Rehabil. Eng.*, vol. 28, no. 2, pp. 461–467, 2020, doi: 10.1109/TNSRE.2020.2965029.

[13] H. Xia, J. Kwon, P. Pathak, J. Ahn, P. B. Shull, and Y. L. Park, 'Design of A Multi-Functional Soft Ankle Exoskeleton for Foot-Drop Prevention, Propulsion Assistance, and Inversion/Eversion Stabilization', presented at the 2020 8th IEEE RAS/EMBS International Conference for Biomedical Robotics and Biomechatronics (BioRob), Dec. 2020, pp. 118–123. doi: 10.1109/BioRob49111.2020.9224420.

[14] J. Bae *et al.*, 'A Lightweight and Efficient Portable Soft Exosuit for Paretic Ankle Assistance in Walking After Stroke', presented at the 2018 IEEE International Conference on Robotics and Automation (ICRA), May 2018, pp. 2820–2827. doi: 10.1109/ICRA.2018.8461046.

[15] P. Malcolm, S. Galle, P. Van den Berghe, and D. De Clercq, 'Exoskeleton assistance symmetry matters: unilateral assistance reduces metabolic cost, but relatively less than bilateral assistance', *J. NeuroEngineering Rehabil.*, vol. 15, no. 1, p. 74, Aug. 2018, doi: 10.1186/s12984-018-0381-z.

[16] X. Jin, Y. Cai, A. Prado, and S. K. Agrawal, 'Effects of exoskeleton weight and inertia on human walking', presented at the 2017 IEEE International Conference on Robotics and Automation (ICRA), Jun. 2017, pp. 1772–1777. doi: 10.1109/ICRA.2017.7989210.

[17] M. A. Normand, J. Lee, H. Su, and J. S. Sulzer, 'The effect of hip exoskeleton weight on kinematics, kinetics, and electromyography during human walking', *J. Biomech.*, vol. 152, p. 111552, May 2023, doi: https://doi.org/10.1016/j.jbiomech.2023.111552.

[18] I. Coifman, R. Kram, and R. Riemer, 'Metabolic power response to added mass on the lower extremities during running', *Appl. Ergon.*, vol. 114, p. 104109, Jan. 2024, doi: https://doi.org/10.1016/j.apergo.2023.104109.

[19] R. C. Browning, J. R. Modica, R. Kram, and A. Goswami, 'The effects of adding mass to the legs on the energetics and biomechanics of walking', *Med Sci Sports Exerc*, vol. 39, no. 3, pp. 515–25, Mar. 2007, doi: 10.1249/mss.0b013e31802b3562.

[20] C. Siviy et al., 'Opportunities and challenges in the development of exoskeletons for locomotor assistance', *Nat. Biomed. Eng.*, vol. 7, no. 4, pp. 456–472, Apr. 2023, doi: 10.1038/s41551-022-00984-1.

[21] G. S. Sawicki, O. N. Beck, I. Kang, and A. J. Young, 'The exoskeleton expansion: improving walking and running economy', *J. NeuroEngineering Rehabil.*, vol. 17, no. 1, p. 25, Feb. 2020, doi: 10.1186/s12984-020-00663-9.

[22] G. M. Gasparri, M. O. Bair, R. P. Libby, and Z. F. Lerner, 'Verification of a Robotic Ankle Exoskeleton Control Scheme for Gait Assistance in Individuals with Cerebral Palsy', presented at the 2018 IEEE/RSJ International Conference on Intelligent Robots and Systems (IROS), Oct. 2018, pp. 4673–4678. doi: 10.1109/IROS.2018.8593904.

[23] B. A. Shafer, S. A. Philius, R. W. Nuckols, J. McCall, A. J. Young, and G. S. Sawicki, 'Neuromechanics and Energetics of Walking With an Ankle Exoskeleton Using Neuromuscular-Model Based Control: A Parameter Study', *Front. Bioeng. Biotechnol.*, vol. Volume 9-2021, 2021, doi: 10.3389/fbioe.2021.615358.

[24] Z. Wang, C. Chen, F. Yang, Y. Liu, G. Li, and X. Wu, 'Real-Time Gait Phase Estimation Based on Neural Network and Assistance Strategy Based on Simulated Muscle Dynamics for an Ankle Exosuit', *IEEE Trans. Med. Robot. Bionics*, vol. 5, no. 1, pp. 100–109, 2023, doi: 10.1109/tmrb.2023.3240284.

[25] H. J. Woltring, 'A Fortran package for generalized, cross-validatory spline smoothing and differentiation', *Adv. Eng. Softw. 1978*, vol. 8, no. 2, pp. 104–113, Apr. 1986, doi: 10.1016/0141-1195(86)90098-7.

[26] B. Michaud and M. Begon, 'ezc3d: An easy C3D file I/O cross-platform solution for C++, Python and MATLAB', *J. Open Source Softw.*, vol. 6, no. 58, p. 2911, 2021, doi: 10.21105/joss.02911.

[27] J. Zhang et al., 'Human-in-the-loop optimization of exoskeleton assistance during walking', *Science*, vol. 356, no. 6344, pp. 1280–1284, Jun. 2017, doi: 10.1126/science.aal5054.

[28] Y. Li, Z. Li, G. Li, T. Wang, J. Zhang, and D. Fu, 'Learning-Based Trajectory Adaption and Neural Network-Based Control of a Soft Exosuit', presented at the 2023 IEEE International Conference on Development and Learning (ICDL), Nov. 2023, pp. 506–511. doi: 10.1109/ICDL55364.2023.10364500.

[29] G. Orekhov and Z. F. Lerner, 'Design and Electromechanical Performance Evaluation of a Powered Parallel-Elastic Ankle Exoskeleton', *IEEE Robot. Autom. Lett.*, vol. 7, no. 3, pp. 8092–8099, 2022, doi: 10.1109/LRA.2022.3185372.

[30] L. M. Mooney, E. J. Rouse, and H. M. Herr, 'Autonomous exoskeleton reduces metabolic cost of human walking', *J. NeuroEngineering Rehabil.*, vol. 11, no. 1, p. 151, Nov. 2014, doi: 10.1186/1743-0003-11-151.

[31] L. Zhao et al., 'Biomechanical Design, Modeling and Control of an Ankle-Exosuit System', in Intelligent Robotics and Applications. Springer Nature Singapore, 2023, pp. 489–502.